\documentclass[conference,letterpaper]{IEEEtran}
\IEEEoverridecommandlockouts
\makeatletter
\newcommand{\linebreakand}{%
  \end{@IEEEauthorhalign}
  \hfill\mbox{}\par
  \mbox{}\hfill\begin{@IEEEauthorhalign}
}
\makeatother
\usepackage[caption=false,font=normalsize,labelfont=sf,textfont=sf]{subfig}
\usepackage{makecell}
\usepackage[T1]{fontenc}
\usepackage{cite}
\usepackage{marvosym}
\usepackage{amsmath,amssymb,amsfonts}

\usepackage{algorithmic}
\usepackage{graphicx}
\usepackage{textcomp}
\usepackage{xcolor}
\usepackage{enumitem}
\usepackage{multirow}      
\usepackage{booktabs}
\usepackage[english]{babel}

\usepackage[table]{xcolor} 
\usepackage{colortbl}     
\usepackage{siunitx}
\usepackage{amsmath}       
\usepackage{threeparttable}
\usepackage{tcolorbox}
\usepackage{fancyhdr}

\definecolor{mylightblue}{RGB}{100,149,237} 
\definecolor{cvprblue}{rgb}{0.21,0.49,0.74}
\definecolor{myyellow}{HTML}{FFCB64}
\definecolor{myteal}{HTML}{62D4C5}
\definecolor{myblue}{HTML}{3266E3}
\definecolor{mypink}{HTML}{FE839C}

\usepackage[pagebackref,breaklinks,colorlinks,allcolors=cvprblue]{hyperref}

\AtBeginDocument{%
  \hypersetup{
    colorlinks=true,
    citecolor=cvprblue,  
    linkcolor=cvprblue,
    urlcolor=cvprblue
  }%
}

\def\BibTeX{{\rm B\kern-.05em{\sc i\kern-.025em b}\kern-.08em
    T\kern-.1667em\lower.7ex\hbox{E}\kern-.125emX}}

\begin{document}

\title{\vspace*{12pt}\textsc{Bridge}: Retrieval-Augmented Spatiotemporal Modeling for Urban Delivery Demand
\vspace{-4pt}}

\author{
\textbf{Yihong Tang}\textsuperscript{1}\quad
\textbf{Tong Nie}\textsuperscript{2}\quad
\textbf{Junlin He}\textsuperscript{2}\quad
\textbf{Qianjun Huang}\textsuperscript{3}\quad
\textbf{Dingyi Zhuang}\textsuperscript{4}\quad
\textbf{Lijun Sun}\textsuperscript{1 \Letter}\\
\textsuperscript{1}McGill University \quad
\textsuperscript{2}The Hong Kong Polytechnic University\quad
\textsuperscript{3}University of Toronto \quad
\textsuperscript{4}MIT\\
\small{
\texttt{\href{mailto:yihong.tang@mail.mcgill.ca}{yihong.tang@mail.mcgill.ca}}, \quad
\texttt{\href{mailto:lijun.sun@mcgill.ca}{lijun.sun@mcgill.ca} (Corresponding)}
}
\vspace{-3.5pt}
}

\maketitle

\maketitle

\pagestyle{fancy}
\fancyhf{}
\fancyhead[C]{\small 2026 IEEE 29th International Conference on Intelligent Transportation Systems (ITSC 2026)}
\fancyfoot[C]{\thepage}
\renewcommand{\headrulewidth}{0pt}

\thispagestyle{fancy}

\begin{abstract}
Forecasting urban delivery demand becomes substantially more challenging when newly added service regions lack historical records. Existing spatiotemporal forecasters effectively model spatial dependence once sufficient node histories are available. Still, they remain parametric and therefore struggle to recover short-term operational dynamics in cold-start regions. Geospatial embeddings help identify where a region is and what function it serves, yet they do not directly reveal how a similar region behaves under a comparable temporal context.
We propose \textsc{Bridge}, a retrieval-augmented spatiotemporal graph framework that combines an inductive contextual graph backbone with a time-aware memory of region-time windows. For each target region, \textsc{Bridge} retrieves future demand patterns from the memory using both regional context and recent dynamics, and refines the backbone forecast through a gated fusion mechanism. To align retrieval with forecasting utility, we further train the retriever with a future-aware objective that favors entries whose future trajectories best match the target. Experiments on four real-world delivery datasets show that \textsc{Bridge} consistently improves over competitive spatiotemporal baselines in both within-city cold-start and cross-city transfer with partial observations. The results show that retrieval augmentation provides a useful operational memory for cold-start urban demand forecasting when parametric graph generalization alone is insufficient.
\end{abstract}


\section{Introduction}
Urban delivery demand forecasting is a core component of last-mile logistics because it directly affects locker deployment, depot planning, courier dispatching, and service reliability~\cite{yannis2006effects,hess2021real,srinivas2022autonomous,nie2025joint}. As e-commerce networks expand, demand estimation is no longer needed only for mature service zones with long operational records. Operators frequently open new delivery regions, add parcel lockers, or expand to suburban districts where little or no demand history is available. In such cases, the forecasting problem is inherently inductive: the model must infer future demand for regions that cannot be treated as fully supervised time series.

\begin{figure}[t]
    \centering
    \includegraphics[width=.95\linewidth]{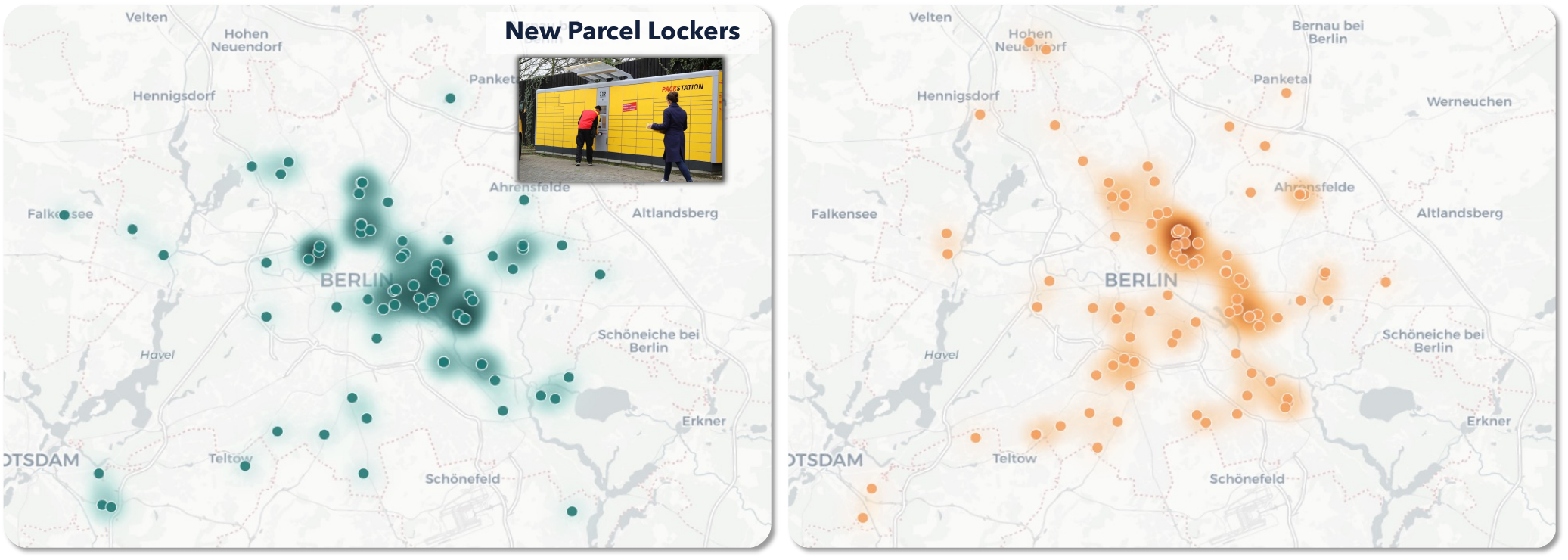}
    \vspace{-10pt}
    \caption{Parcel-locker expansion in Berlin, 2025. The left and right panels show lockers deployed in the first and second halves of the year, respectively. New facilities may be important but initially lack local demand histories.}
    \label{fig:intro}
    \vspace{-18pt}
\end{figure}

Real-world delivery networks are dynamic by nature. Operators expand parcel-locker systems, open new service areas, or reorganize regions, and newly introduced locations often have little or no historical demand at deployment time. Even in existing cities, some regions may only be partially observed due to recent rollout or limited data collection. As a result, forecasting must operate under partial observability: the model needs to estimate future demand for regions with weak or missing local histories, rather than simply extrapolating well-observed time series. Figure~\ref{fig:intro} illustrates a representative expansion scenario, where newly deployed lockers are geographically and operationally important but lack sufficient historical records to support conventional graph-based forecasting.

Graph-based spatiotemporal models are a natural foundation for this problem because delivery demand is spatially correlated across nearby or functionally similar regions \cite{DCRNN,GWNET,MTGNN,cini2023graph,tang2022domain}. By propagating information over a graph, these models can transfer signals from well-observed regions to those with limited data. However, they remain fully parametric: when a region has little history or exhibits a regime that is underrepresented in training, the model must rely solely on its learned parameters to extrapolate future demand. Even when enhanced with geospatial embeddings from large language models (LLMs) that encode static regional semantics \cite{gurnee2023language,manvi2023geollm,he2025geolocation}, the predictor still lacks explicit access to how similar regions evolved under comparable short-term conditions.

To address this limitation, we introduce retrieval augmentation into spatiotemporal forecasting. Instead of requiring the graph networks to implicitly memorize diverse demand dynamics, we maintain a memory of historical regional delivery patterns and their corresponding future demands. For a target region at the current time, the model retrieves past windows that are similar in regional context and recent dynamics, and uses their stored futures as a prior. This way, retrieval directly supplies candidate future patterns, complementing the graph model when local observations are sparse or missing.

Based on this idea, we propose \textsc{Bridge}, short for \underline{B}ank-\underline{R}etrieval \underline{I}nductive \underline{D}emand \underline{G}raph l\underline{E}arning. \textsc{Bridge} combines an inductive contextual graph backbone with time-aware retrieval of future priors, a gated fusion mechanism in prediction space, and a future-aware retriever objective aligned with forecasting utility. The design is intentionally targeted at the two settings above: single-city cold-start forecasting and cross-city transfer with partial target observations.

The contributions of this paper are threefold:
\begin{itemize}[leftmargin=*]
    \item We study delivery demand forecasting under partial observability, where some regions have little or no historical demand at prediction time, and frame it as a retrieval-augmented spatiotemporal learning problem.
    \item We propose \textsc{Bridge}, which integrates an inductive contextual graph backbone with time-aware retrieval of possible future demand patterns and gated forecast refinement, together with a future-aware training objective for the retriever.
    \item We conduct comprehensive experiments on four real-world delivery datasets. Results show that our method consistently outperforms powerful baselines in both within-city cold-start and cross-city transfer scenarios.
\end{itemize}

\section{Related Work}

\paragraph{Delivery Demand Forecasting}
Prior work studies demand estimation, route prediction, and platform-level forecasting using statistical and deep learning models \cite{hess2021real,liang2023poisson,wen2024survey}. Most methods assume sufficient historical records for all regions. In practice, however, delivery networks evolve continuously, and newly deployed or sparsely monitored regions often lack adequate history, leading to structural missingness at the regional level. Recent work such as \textsc{Impel}~\cite{nie2025joint} explores inductive training strategies to improve generalization to unseen regions, but these approaches remain fully parametric and must extrapolate demand patterns solely from learned weights.

\paragraph{Spatiotemporal Graph Modeling}
Spatiotemporal graph neural networks (STGNNs) model correlated regional demand via spatial propagation and temporal dynamics \cite{DCRNN,GWNET,MTGNN,MPGRU}. Extensions to graph-based imputation and kriging address partially observed nodes \cite{IGNNK,GRIN,wei2024inductive}. Compared with imputation, our setting requires forecasting future demand for all regions while some are cold-start at prediction time, and considers deployment across cities under partial observability.

\paragraph{Geospatial Representations and Retrieval}
Large language models (LLMs) have been used to derive transferable geospatial embeddings from textual and urban metadata \cite{gurnee2023language,manvi2023geollm}. These embeddings provide static semantic priors but do not capture short-term demand dynamics. Retrieval-augmented learning reuses non-parametric historical examples \cite{lewis2020retrieval} and offers a complementary mechanism. In this work, we retrieve historical region-time windows and their future demand patterns to support cold-start delivery forecasting.

\section{Problem Formulation}

We consider city-wide delivery demand aggregated to a set $R$ of regions of interest (ROIs).
Let $r_i \in R$ denote the $i$-th region.
The delivery demand $\mathbf{x}_t^i$ is defined as the number of orders whose pickup locations fall into region $r_i$ during the $t$-th time interval.
Given a historical window of length $W$ and a forecasting horizon of length $H$, we denote the historical and future delivery demands of region $i$ as
\begin{equation}
\mathbf{x}_{t-W+1:t}^{i} \in \mathbb{R}^{W}, 
\quad
\mathbf{y}_{t+1:t+H}^{i} \in \mathbb{R}^{H}.
\end{equation}
Stacking all $N$ regions yields:
\begin{equation}
\mathbf{X}_{t-W+1:t} \in \mathbb{R}^{W \times N}, 
\quad
\mathbf{Y}_{t+1:t+H} \in \mathbb{R}^{H \times N}.
\end{equation}
Not all regions are assumed to have observable histories.
Let $R^\mathcal{O}_t \subseteq R=\{r_1,\dots,r_N\}$ denote the set of regions whose historical demand is observable at time $t$, and let $R^\mathcal{U}_t = R\setminus R^\mathcal{O}_t$ denote the set of unobserved (cold-start) regions.
We define a binary observation mask $\mathbf{m}_t \in \{0,1\}^{N}$ such that $m_t^i = 1$ if $r_i \in R^\mathcal{O}_t$ and $m_t^i = 0$ otherwise.
Then the input at time $t$ is:
\begin{equation}
\mathbf{X}^{\mathcal{O}}_{t-W+1:t}
=
\{ \mathbf{x}_{t-W+1:t}^{i} \mid r_i \in R^\mathcal{O}_t \}.
\end{equation}
Each region $r_i$ is associated with a vector 
$\mathbf{c}_i \in \mathbb{R}^{d_c}$ that could summarize its geographic and demographic contexts. These regional contexts are constructed from region-level metadata and are assumed to be time-invariant. Let $\mathbf{C} = [\mathbf{c}_1,\dots,\mathbf{c}_N]^\top 
\in \mathbb{R}^{N \times d_c}$ denote the contexts of all regions, the forecasting objective is to learn a predictor $f_{\theta}$:
\begin{equation}
\widehat{\mathbf{Y}}_{t+1:t+H}
=
f_{\theta}\!\left(
\mathbf{X}^{\mathcal{O}}_{t-W+1:t},
\mathbf{C},
\mathbf{m}_t
\right),
\end{equation}
that estimates future demand for all regions, including those in $R^\mathcal{U}_t$ whose histories are unavailable.

\begin{figure*}[t]
    \centering
    \includegraphics[width=.8\textwidth]{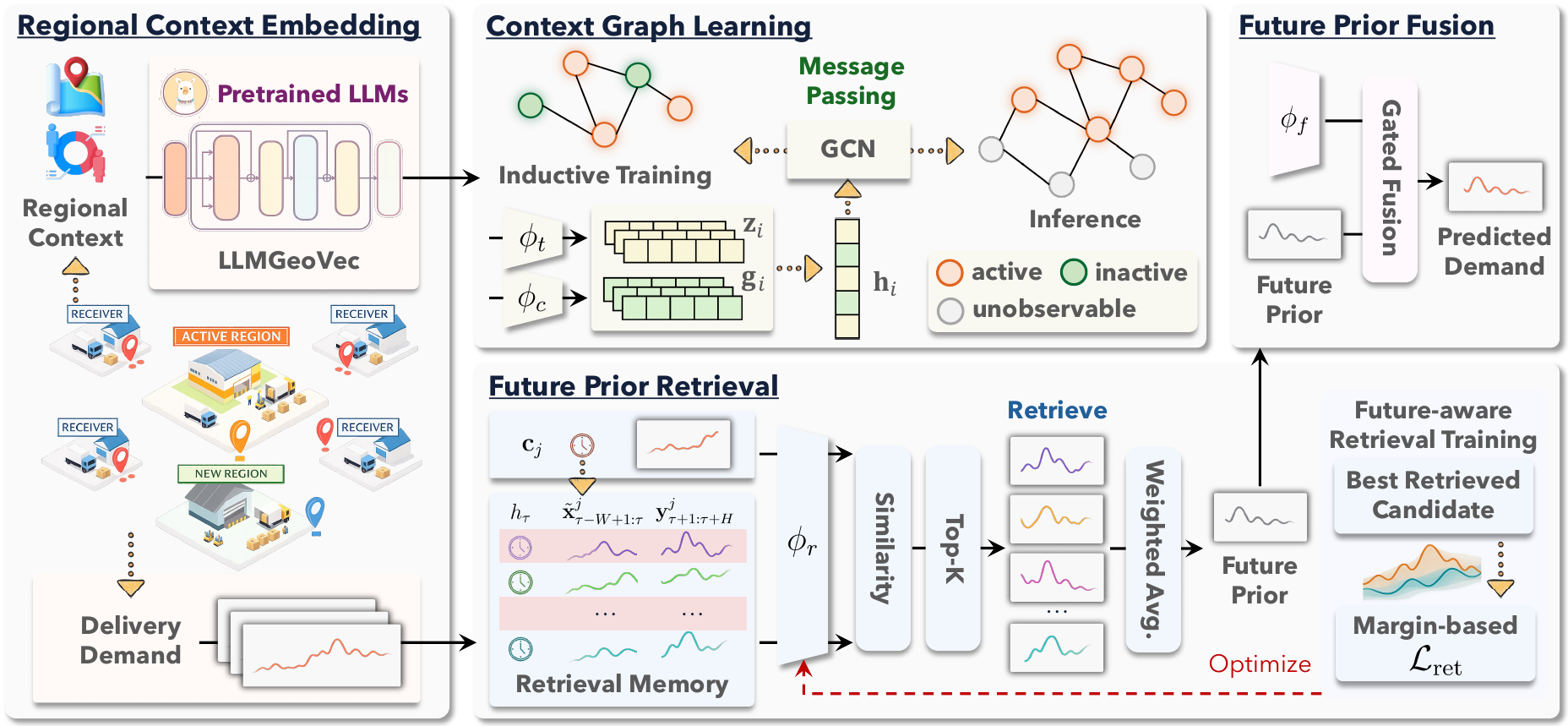}
    \vspace{-9pt}
    \caption{Overview of \textsc{Bridge}. Region contexts and historical demand are encoded for inductive graph forecasting; a time-aware memory retrieves future priors from similar region-time patterns; the retrieved priors are then fused with the graph forecast for final prediction.}
    \label{fig:method}
    \vspace{-14pt}
\end{figure*}

\section{Method}
Figure~\ref{fig:method} gives an overview of the \textsc{Bridge} framework.

\subsection{Regional Context Embedding}
The first design question is how to represent a region. In our task, cold-start regions provide little or no demand history, so the model cannot rely on time series alone to identify their roles in the urban system. We therefore require a region descriptor that is available for both observed and unobserved regions and remains comparable across cities. In practice, the regional context $\mathbf{c}_i$ is derived from geographic and demographic attributes collected from OpenStreetMap and related public urban metadata. These structured descriptions are encoded using \textsc{LLMGeoVec}~\cite{he2025geolocation}, which maps textual region-level information into a dense vector representation, providing a unified description of region function, land-use, and surrounding urban context compared with manually engineered covariates. Such static contextual information complements dynamic demand signals and provides a transferable prior for regions with limited or missing histories. To obtain a compact representation for downstream retrieval and graph construction, we map $\mathbf{c}_i$ into a $d_g$-dimensional latent space:
\begin{equation}
\mathbf{g}_i = \phi_c\!\left(\mathbf{c}_i\right) = \mathbf{W}_c \mathbf{c}_i \in \mathbb{R}^{d_g}.
\end{equation}
The projection $\phi_c(\cdot)$ reduces the embedding dimension and is optimized with the forecasting objective.

\subsection{Contextual Graph Learning}

To enable forecasting under partial observability, the model must propagate information from regions with available histories to those without.
At each time $t$, only a subset of regions $R_t^{\mathcal O}$ has observed demand histories. Training directly on a fixed observed set may overfit transductively to seen nodes, while deployment requires inductive generalization to unseen or newly revealed regions.
Therefore, at each training instance, we randomly sample a subset $R_t^{\mathrm{act}} \subseteq R_t^{\mathcal O}$ as the \emph{active} regions whose historical inputs are retained, and treat the remaining regions $R_t^{\mathrm{inact}} = R_t^{\mathcal O} \setminus R_t^{\mathrm{act}}$ as \emph{inactive}, whose histories are masked to simulate cold-start conditions.
During training, we construct an inductive graph over the observable region set $R^\mathcal{O}_t$. At inference time, we build the graph over all regions $R$, where edges are defined by contextual similarity computed from $\{\mathbf{g}_i\}$. Therefore, the adjacency matrix at time $t$ is:

\vspace{-2.5mm}
\begin{equation}
\mathbf{A}_t
=
\operatorname{softmax}\!\left(
\operatorname{GeLU}(\mathbf{G}_t \mathbf{G}_t^\top)
\right),
\label{eq:sem_graph}
\end{equation}
\vspace{-6mm}

\noindent where $\mathbf{G}_t$ stacks the contextual representations of regions participating in the graph at time $t$. During training, this corresponds to regions in $R_t^{\mathcal O}$, while at inference time it includes all regions in $R$.
The graph structure is independent of demand availability, allowing both active and inactive regions to participate in message passing.
For each region $r_i$, we extract a temporal representation from its historical demand:
\vspace{-1mm}
\begin{equation}
\mathbf{z}_i = \phi_t\!\left(\widetilde{\mathbf{x}}_{t-W+1:t}^{i}\right) \in \mathbb{R}^{d_z},
\end{equation}
\vspace{-5mm}

\noindent where $\phi_t(\cdot)$ is a temporal encoder as a linear projection over the history window.
The masked history is defined as:
\vspace{-1mm}
\[
\widetilde{\mathbf{x}}_{t-W+1:t}^{i} =
\begin{cases}
\mathbf{x}_{t-W+1:t}^{i}, & r_i \in R_t^{\mathrm{act}},\\
\mathbf{0}, & r_i \in R_t^{\mathrm{inact}}.
\end{cases}
\]
\vspace{-3mm}

\noindent Thus, inactive regions remain in the graph but do not contribute to temporal signals.
The initial node representation is formed by concatenating temporal and contextual components:
$\mathbf{h}_i^{(0)} = [\mathbf{z}_i \, \| \, \mathbf{g}_i].$
Message passing~\cite{kipf2016semi} is then performed as:
\vspace{-0.5mm}
\begin{equation}
\mathbf{H}^{(\ell+1)}
=
\operatorname{GCN}\!\left(\mathbf{H}^{(\ell)}, \mathbf{A}_t\right)
+
\mathbf{H}^{(\ell)},
\, \ell = 0,\dots,L-1,
\end{equation}
\vspace{-3.5mm}

\noindent where $\mathbf{H}^{(0)} = [\mathbf{h}_1^{(0)},\dots,\mathbf{h}_{|R^\mathcal{O}_t|}^{(0)}]^\top$. This contextual backbone enables information propagation from active to inactive regions, encouraging the model to learn inductive transfer across nodes while maintaining a shared parametric predictor.

\subsection{Learning to Retrieve Future Priors}

Cold-start demand forecasting is often governed by short-term operational dynamics, such as lunch peaks, evening surges, and time-of-day specific patterns. These patterns are highly regular in the sense that similar regional contexts can exhibit similar dynamics under comparable temporal conditions, yet they are difficult to infer from parameters alone when historical observations are missing. We therefore equip the model with a time-aware retrieval mechanism that provides a future prior by reusing region-time patterns stored in memory.

\subsubsection{Memory Construction and Representation}
Let $\mathcal{B}$ denote the retrieval memory bank.
Each entry corresponds to a region-time window of total length $W+H$ ending at time $\tau$.
For a region $r_j$ and a window ending at $\tau$, we store

\vspace{-2.5mm}
\begin{equation}
\mathbf{b}_{\tau,j} =
\left(
\mathbf{c}_j, \,
\widetilde{\mathbf{x}}_{\tau-W+1:\tau}^{j}, \,
\mathbf{y}_{\tau+1:\tau+H}^{j},\,
h_{\tau}
\right),
\end{equation}
\vspace{-4mm}

\noindent where $\mathbf{c}_j$ is the regional context,
$\widetilde{\mathbf{x}}_{\tau-W+1:\tau}^{j}$ is the historical demand segment used as the retrieval key, $\mathbf{y}_{\tau+1:\tau+H}^{j}$ is the corresponding future demand, and $h_\tau$ denotes the time-of-day indicator of the window. This design ensures that retrieval reflects both regional context and short-term dynamics.
Given a region $r_i$ at time $t$, we form a query embedding:

\vspace{-3mm}
\begin{equation}
\mathbf{q}_{i}
=
\phi_r\!\left(
\mathbf{c}_i,\,
\widetilde{\mathbf{x}}_{t-W+1:t}^{i},\,
h_t
\right).
\end{equation}
\vspace{-4mm}

\noindent For regions without observable histories, $\widetilde{\mathbf{x}}_{t-W+1:t}^{i}$ is zero-masked, so retrieval remains applicable under cold-start conditions. For each memory entry, a key embedding is:
\begin{equation}
\mathbf{k}_{\tau,j}
=
\phi_r\!\left(
\mathbf{c}_j,\,
\widetilde{\mathbf{x}}_{\tau-W+1:\tau}^{j},\,
h_{\tau}
\right),
\end{equation}
where $\phi_r(\cdot)$ is a retrieval encoder. 
For any context feature $\mathbf{c}$, temporal feature $\mathbf{x}$, and time indicator $h$, we first compute:
\begin{equation}
\mathbf{e}^{c}=\mathbf{W}_{r}\mathbf{c}, \quad
\mathbf{e}^{x}=\mathbf{W}_{x}[\mathbf{x}\,\|\,\mathbf{e}^{h}],
\end{equation}
where $\mathbf{e}^{h}$ denotes the embedding encoding the time of day $h$. The retrieval representation is then normalized by:
\begin{equation}
\phi_r(\mathbf{c},\mathbf{x},h)
=
\frac{
\psi_r\!\left([\mathbf{e}^{c}\,\|\,\mathbf{e}^{x}]\right)
}{
\left\|\psi_r\!\left([\mathbf{e}^{c}\,\|\,\mathbf{e}^{x}]\right)\right\|_2
},
\end{equation}
where $\psi_r(\cdot)$ is a fusion MLP network. Retrieval is thus performed in a joint context-and-dynamics space: regional contexts identify functionally similar regions, while short-term histories and time information identify temporal dynamics.

\subsubsection{Time-aware Retrieval}

Since demand patterns are strongly conditioned on time-of-day patterns, we restrict retrieval to memory entries observed under the same time window as the query, which also reduces computational overhead: $\Omega_t = 
\left\{
(\tau,j) \in \mathcal{B} 
\;:\;
h_\tau = h_t
\right\}.$ Similarity is computed as:

\vspace{-3mm}
\begin{equation}
s_{i,\tau,j}
=
\mathbf{q}_i^\top \mathbf{k}_{\tau,j},
\end{equation}
\vspace{-5mm}

\noindent followed by top-$\mathrm{K}$ temperature-scaled normalization:
\vspace{-1mm}
\begin{equation}
\alpha_{i,\tau,j}
=
\frac{
\exp(s_{i,\tau,j}/T_r)
}{
\sum_{(\tau',j') \in \operatorname{TopK}(i)}
\exp(s_{i,\tau',j'}/T_r)
}.
\end{equation}
The retrieved future prior for region $r_i$ is defined as the weighted average of the corresponding stored futures:
\vspace{-1mm}
\begin{equation}
\mathbf{p}_i
=
\sum_{(\tau,j) \in \operatorname{TopK}(i)}
\alpha_{i,\tau,j}
\,
\mathbf{y}_{\tau+1:\tau+H}^{j}.
\label{eq:bank_prior_rewrite}
\end{equation}
\vspace{-3mm}

\subsubsection{Future-aware Retrieval Training}

Since the retrieved trajectories are used as forecasting priors, retrieval quality should be measured by alignment with the target future. We therefore introduce a future-aware auxiliary objective that directly optimizes the retriever with respect to this criterion. For each query, we first keep the top-$\mathrm{K}$ candidates based on encoder similarity, and select the candidate whose future pattern is closest to the query's future:

\vspace{-4mm}
\begin{equation}
(\tau^\star,j^\star)
=
\arg\min_{(\tau,j)\in \operatorname{TopK}(i)}
\left\|
\mathbf{y}_{t+1:t+H}^{i}
-
\mathbf{y}_{\tau+1:\tau+H}^{j}
\right\|_2.
\end{equation}
\vspace{-3mm}

\noindent We then encourage the query embedding to align with this future-nearest key via a margin-based loss:

\vspace{-3mm}
\begin{equation}
\mathcal{L}_{\text{ret}}
=
\frac{1}{N_t}
\sum_{i=1}^{N_t}
\left(
1 - \mathbf{q}_i^\top \mathbf{k}_{\tau^\star,j^\star}
\right),
\end{equation}
\vspace{-3mm}

\noindent where $N_t$ denotes the number of regions in the current forecasting instance. This objective connects retrieval learning with forecasting utility, pushing the retriever toward entries that provide accurate future priors.

\subsection{Future Prior Fusion}

After $L$ layers of contextual message passing, each region $r_i$ obtains a latent representation $\mathbf{h}_i^{(L)}$.
A forecasting head $\phi_f(\cdot)$ maps this representation to a backbone prediction:
\vspace{-1mm}
\begin{equation}
\widetilde{\mathbf{y}}^{i}_{t+1:t+H}
=
\phi_f\!\left(\mathbf{h}_i^{(L)}\right)
\in
\mathbb{R}^{H}.
\end{equation}
\vspace{-4mm}

The retrieved future prior $\mathbf{p}_i$ serves as an additive correction.
We first align it to the prediction space:

\vspace{-3mm}
\begin{equation}
\widetilde{\mathbf{p}}_i
=
\mathbf{W}_p \mathbf{p}_i.
\end{equation}
\vspace{-5mm}

A gating vector conditioned on both signals is computed as
\begin{equation}
\mathbf{u}_i
=
\sigma\!\left(
\mathbf{W}_g
\left[
\widetilde{\mathbf{y}}^{i}_{t+1:t+H}
\,\|
\,
\widetilde{\mathbf{p}}_i
\right]
\right),
\end{equation}
where $\sigma(\cdot)$ denotes the sigmoid function.

The final prediction for region $r_i$ is
\begin{equation}
\widehat{\mathbf{y}}^i_{t+1:t+H}
=
\widetilde{\mathbf{y}}^{i}_{t+1:t+H}
+
\beta
\,
\mathbf{u}_i
\odot
\widetilde{\mathbf{p}}_i,
\label{eq:fusion_rewrite}
\end{equation}
where $\beta$ is a learnable scalar initialized at zero and $\odot$ denotes element-wise multiplication.
This design preserves the graph backbone as the primary predictor while allowing retrieval to provide instance-specific adjustments when beneficial.

\par\vspace*{1pt}

Stacking over all regions yields $\widehat{\mathbf{Y}}_{t+1:t+H}$. At inference time, for regions without observable histories, temporal encodings are zero-masked, and prediction relies on contextual graph propagation and retrieved future priors.

\subsection{Training Objective}

The model is trained with a masked forecasting loss together with the future-aware retrieval objective.
Since predictions are defined at the region level as 
$\widehat{\mathbf{y}}_{t+1:t+H,i} \in \mathbb{R}^{H}$,
we use an $\ell_1$ loss over the forecasting horizon:
\vspace{-1.5mm}
\begin{equation}
\mathcal{L}_{\text{pred}}
=
\frac{1}{|R_t^{\mathcal O}| \cdot H}
\sum_{r_i \in R_t^{\mathcal O}}
\left\|
\widehat{\mathbf{y}}_{t+1:t+H}^i
-
\mathbf{y}_{t+1:t+H}^{i}
\right\|_{1}.
\end{equation}
\vspace{-3.5mm}

Together with the retrieval loss $\mathcal{L}_{\text{ret}}$, the overall objective is

\vspace{-3.5mm}
\begin{equation}
\mathcal{L}
=
\mathcal{L}_{\text{pred}}
+
\lambda_{\text{ret}}
\mathcal{L}_{\text{ret}},
\label{eq:loss_rewrite}
\end{equation}
\vspace{-5mm}

\noindent where $\lambda_{\text{ret}}$ controls the strength of retrieval supervision.

\section{Experiments}

\subsection{Datasets and Experimental Settings}

We evaluate on four processed delivery datasets~\cite{wu2023lade}: Shanghai (SH), Hangzhou (HZ), Yantai (YT), and Chongqing (CQ). Each sample contains a 24-step history window and a 24-step prediction horizon. Table~\ref{tab:data_stats} summarizes the number of regions and data splits. All cities contain around 30 regions, which makes the benchmark suitable for studying node-level cold-start forecasting under sparse supervision.
\vspace{-8pt}

\begin{table}[h]
\caption{Dataset statistics used in this paper.}
\vspace{-5pt}
\label{tab:data_stats}
\centering
\small
\renewcommand{\arraystretch}{.8}
\resizebox{.7\columnwidth}{!}{
\begin{tabular}{lcccc}
\toprule
Cities & \#Nodes & \#Train & \#Val & \#Test \\
\midrule
Shanghai & 30 & 2540 & 847 & 847 \\
Hangzhou & 31 & 2617 & 872 & 872 \\
Yantai & 30 & 2616 & 872 & 872 \\
Chongqing & 30 & 2617 & 872 & 873 \\
\bottomrule
\end{tabular}
}
\vspace{-5pt}
\end{table}

\begin{table*}[t]
\centering
\caption{Cross-city transfer results. Lower MAE/RMSE is better; higher $R^2$ is better.}
\vspace{-5pt}
\label{tab:transfer_all_pairs}
\renewcommand{\arraystretch}{1.05}
\setlength{\tabcolsep}{2.5pt}
\resizebox{\linewidth}{!}{
\begin{tabular}{l|
S S S|
S S S|
S S S|
S S S|
S S S}
\toprule

\multirow{2}{*}{Source $\mapsto$ Target}
& \multicolumn{3}{c|}{\textbf{\textsc{Bridge} (Ours)}}
& \multicolumn{3}{c|}{\textbf{\textsc{Impel}}}
& \multicolumn{3}{c|}{\textbf{MTGNN}}
& \multicolumn{3}{c|}{\textbf{IGNNK}}
& \multicolumn{3}{c}{\textbf{STGCN}} \\

\cmidrule(lr){2-4}
\cmidrule(lr){5-7}
\cmidrule(lr){8-10}
\cmidrule(lr){11-13}
\cmidrule(lr){14-16}

& {MAE$\downarrow$} & {RMSE$\downarrow$} & {R$^2\uparrow$}
& {MAE$\downarrow$} & {RMSE$\downarrow$} & {R$^2\uparrow$}
& {MAE$\downarrow$} & {RMSE$\downarrow$} & {R$^2\uparrow$}
& {MAE$\downarrow$} & {RMSE$\downarrow$} & {R$^2\uparrow$}
& {MAE$\downarrow$} & {RMSE$\downarrow$} & {R$^2\uparrow$} \\

\midrule

Chongqing $\mapsto$ Hangzhou 
& \cellcolor{orange!15}{\textbf{5.18}} 
& \cellcolor{orange!15}{\textbf{9.75}} 
& \cellcolor{orange!15}{\textbf{0.902}}
& 5.37 & 10.11 & 0.894
& 6.92 & 14.70 & 0.775
& 7.45 & 15.27 & 0.756
& 7.33 & 15.14 & 0.758 \\

Chongqing $\mapsto$ Shanghai  
& \cellcolor{orange!15}{\textbf{4.00}} 
& \cellcolor{orange!15}{\textbf{6.29}} 
& \cellcolor{orange!15}{\textbf{0.904}}
& 4.28 & 6.60 & 0.894
& 5.11 & 9.42 & 0.803
& 5.24 & 9.82 & 0.786
& 5.27 & 9.71 & 0.790 \\

Chongqing $\mapsto$ Yantai   
& \cellcolor{orange!15}{\textbf{2.92}} 
& \cellcolor{orange!15}{\textbf{4.79}} 
& \cellcolor{orange!15}{\textbf{0.945}}
& 2.99 & 4.86 & 0.943
& 3.44 & 6.50 & 0.898
& 3.38 & 6.20 & 0.906
& 3.73 & 6.98 & 0.885 \\

Hangzhou $\mapsto$ Chongqing 
& \cellcolor{orange!15}{\textbf{3.64}} 
& \cellcolor{orange!15}{\textbf{6.06}} 
& \cellcolor{orange!15}{\textbf{0.935}}
& 3.86 & 6.37 & 0.928
& 4.79 & 8.93 & 0.859
& 4.89 & 8.85 & 0.861
& 4.92 & 9.06 & 0.854 \\

Hangzhou $\mapsto$ Shanghai  
& \cellcolor{orange!15}{\textbf{3.10}} 
& \cellcolor{orange!15}{\textbf{4.84}} 
& \cellcolor{orange!15}{\textbf{0.926}}
& 3.35 & 5.23 & 0.913
& 4.12 & 7.56 & 0.819
& 4.02 & 7.29 & 0.831
& 4.35 & 7.91 & 0.808 \\

Hangzhou $\mapsto$ Yantai    
& \cellcolor{orange!15}{\textbf{2.27}} 
& \cellcolor{orange!15}{\textbf{3.67}} 
& \cellcolor{orange!15}{\textbf{0.939}}
& 2.43 & 3.89 & 0.931
& 2.87 & 5.01 & 0.885
& 2.89 & 5.10 & 0.881
& 3.14 & 5.47 & 0.863 \\

Shanghai $\mapsto$ Chongqing 
& \cellcolor{orange!15}{\textbf{2.95}} 
& \cellcolor{orange!15}{\textbf{4.75}} 
& \cellcolor{orange!15}{\textbf{0.949}}
& 3.09 & 5.02 & 0.943
& 3.45 & 6.16 & 0.913
& 3.35 & 5.98 & 0.918
& 3.63 & 6.43 & 0.904 \\

Shanghai $\mapsto$ Hangzhou  
& \cellcolor{orange!15}{\textbf{5.24}} 
& \cellcolor{orange!15}{\textbf{9.15}} 
& \cellcolor{orange!15}{\textbf{0.905}}
& 5.54 & 9.50 & 0.896
& 7.14 & 14.29 & 0.773
& 7.13 & 14.52 & 0.765
& 8.02 & 15.44 & 0.740 \\

Shanghai $\mapsto$ Yantai    
& \cellcolor{orange!15}{\textbf{2.79}} 
& \cellcolor{orange!15}{\textbf{4.51}} 
& \cellcolor{orange!15}{\textbf{0.952}}
& 2.82 & 4.57 & 0.951
& 3.38 & 6.26 & 0.907
& 3.38 & 6.19 & 0.909
& 3.68 & 6.80 & 0.890 \\

Yantai $\mapsto$ Chongqing   
& \cellcolor{orange!15}{\textbf{3.64}} 
& \cellcolor{orange!15}{\textbf{6.01}} 
& \cellcolor{orange!15}{\textbf{0.940}}
& 3.71 & 6.10 & 0.938
& 4.53 & 8.29 & 0.885
& 4.25 & 7.71 & 0.901
& 4.62 & 8.52 & 0.879 \\

Yantai $\mapsto$ Hangzhou    
& \cellcolor{orange!15}{\textbf{5.33}} 
& \cellcolor{orange!15}{\textbf{9.34}} 
& \cellcolor{orange!15}{\textbf{0.901}}
& 5.58 & 9.68 & 0.892
& 7.08 & 14.32 & 0.769
& 7.21 & 14.58 & 0.760
& 7.64 & 14.89 & 0.751 \\

Yantai $\mapsto$ Shanghai    
& \cellcolor{orange!15}{\textbf{3.70}} 
& \cellcolor{orange!15}{\textbf{5.76}} 
& \cellcolor{orange!15}{\textbf{0.918}}
& 3.83 & 5.99 & 0.912
& 4.92 & 9.17 & 0.792
& 4.68 & 8.78 & 0.810
& 4.91 & 9.22 & 0.790 \\

\bottomrule
\end{tabular}
}
\vspace{-15pt}

\end{table*}

We study two settings. In \emph{single-city cold-start forecasting}, 10 regions are held out as cold-start targets, and an additional 6 observable regions are randomly masked in each training batch to simulate inductive deployment. In \emph{cross-city transfer with partial target observations}, a model is pretrained on a source city and then directly evaluated on a target city where 10 regions are treated as unobserved at test time. Following the current implementation, training in this setting uses the source-city graph only, while inference is performed on the full target graph with unavailable target histories masked out.

The backbone architecture uses a node embedding dimension of 32, a hidden dimension of 64, three MLP blocks, and one message-passing layer. Retrieval augmentation is enabled with time-aware candidate filtering and a source-memory bank built from source training windows. In the reported runs, the main retrieval hyper-parameters are $K= 8$, retriever dimension is $128$, and $\lambda_{\text{ret}} =0.2$.
We compare \textsc{Bridge} against its graph-only backbone and a broader family of spatiotemporal baselines, including DCRNN, STGCN, Graph WaveNet, MTGNN, SATCN, MPGRU, GRIN, IGNNK, and \textsc{Impel}~\cite{DCRNN,STGCN,GWNET,MTGNN,MPGRU,GRIN,IGNNK,nie2025joint}. We mainly report MAE and RMSE for the single-city setting, and MAE, RMSE, and $R^2$ for cross-city transfer with partial target observations.

\subsection{Cross-City Transfer with Partial Target Observations}

\begin{figure}[h!]
    \centering
    \vspace{-12pt}
    \includegraphics[width=\columnwidth]{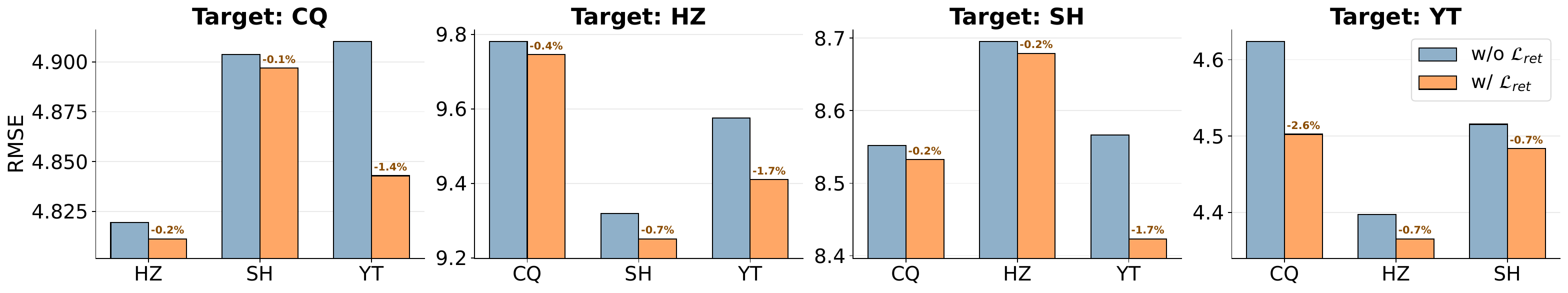}
    \vspace{-23pt}
    \caption{Ablation of the $\mathcal{L}_{\mathrm{ret}}$ in the cross-city transfer setting.}
    \label{fig:ablation}
\vspace{-7pt}
\end{figure}

Table~\ref{tab:transfer_all_pairs} summarizes the cross-city transfer results over all 12 ordered source-target pairs. \textsc{Bridge} achieves the best MAE, RMSE, and $R^2$ in every pair, outperforming the graph-only backbone as well as MTGNN, IGNNK, and STGCN. The gains over the graph-only backbone are sometimes modest, but they are consistent: the average MAE decreases from 3.904 to 3.730, and the average RMSE decreases from 6.494 to 6.229 across all transfer pairs.
Two patterns are worth highlighting. First, transfer between the larger and more active cities remains challenging, yet retrieval still yields reliable improvements, such as Chongqing $\mapsto$ Hangzhou and Shanghai $\mapsto$ Hangzhou. Second, when the target city is relatively easier to predict, such as Yantai, \textsc{Bridge} still provides measurable gains, indicating that retrieval does not only help under severe mismatch but also acts as a useful prior under moderate transfer shifts. Overall, these results support the main motivation of the paper: cross-city cold-start forecasting benefits from a non-parametric future prior that complements graph generalization. Figure~\ref{fig:ablation} shows that the $\mathcal{L}_\mathrm{ret}$ consistently reduces RMSE for every evaluated source-target pair, indicating that explicitly training the retriever toward future-useful memories improves cross-city transfer robustness.

\subsection{Single-City Cold-Start Results}

\vspace{-1pt}
\begin{table}[htbp]
\vspace{-2pt}
\caption{Experimental results of spatiotemporal delivery demand joint estimation and prediction in package delivery datasets.}
\vspace{-4pt}
\label{tab:result_main_lade}
\centering
\begin{small}
    \renewcommand{\multirowsetup}{\centering}
    \setlength{\tabcolsep}{3pt}
    \renewcommand{\arraystretch}{1}
    \resizebox{\columnwidth}{!}{
    \begin{tabular}{l|cc|cc|cc|cc}
    \toprule
     \multirow{2}{*}{Models} & \multicolumn{2}{c|}{\textbf{Shanghai}} & \multicolumn{2}{c|}{\textbf{Hangzhou}} & \multicolumn{2}{c|}{\textbf{Chongqing}} & \multicolumn{2}{c}{\textbf{Yantai}} \\
    \cmidrule(lr){2-9} 
    & MAE & RMSE & MAE & RMSE & MAE & RMSE & MAE & RMSE  \\
    \midrule
    HA  & 6.96 & 16.41 & 8.94 & 20.56 & 4.00 & 8.85 & 3.82 & 8.42\\
    DCRNN \cite{DCRNN} & 5.65 & 11.86 & 7.33 & 14.59 & 3.53 & 6.15 & 3.21 & 6.17 \\
    STGCN \cite{STGCN} & 5.07 & 11.62 & 6.38 & 14.26 & 2.99 & 6.00 & 2.80 & 5.79 \\
    GWNET \cite{GWNET} & 5.22 & 11.67 & 7.99 & 15.90 & 3.06 & 6.03 & 2.93 & 6.01\\
    MTGNN \cite{MTGNN} & 5.09 & 11.56 & 6.23 & 13.89 & 2.97 & 5.91 & 2.73 & 5.70 \\
    IGNNK \cite{IGNNK} & 5.22 & 11.50 & 7.25 & 15.06 & 3.22 & 6.06 & 2.97 & 5.93 \\
    SATCN \cite{SATCN} & 4.75 & 9.38  & 7.64 & 14.77  & 3.04 & 5.27 & 2.83 & 5.02 \\
    MPGRU \cite{MPGRU} & 6.30 & 13.43 & 7.95 & 16.03 & 3.91 & 7.60 & 3.58 & 7.45 \\
    GRIN \cite{GRIN}& 5.08 & 11.64 & 6.30 & 14.56 & 3.05  & 6.08 & 2.86 & 6.02 \\
    \textsc{Impel} \cite{nie2025joint} & 3.76 & 7.93  & 4.52 & 9.90 & 2.47 & 4.92 & 2.23 & 4.18 \\
    \midrule
    \textbf{\textsc{Bridge}} & \cellcolor{orange!15}{\textbf{3.71}} & \cellcolor{orange!15}{\textbf{7.67}}  & \cellcolor{orange!15}{\textbf{4.30}} & \cellcolor{orange!15}{\textbf{9.46}} & \cellcolor{orange!15}{\textbf{2.41}} & \cellcolor{orange!15}{\textbf{4.71}} & \cellcolor{orange!15}{\textbf{2.19}} & \cellcolor{orange!15}{\textbf{4.08}} \\
    \bottomrule
    \end{tabular}}
\end{small}
\vspace{-15pt}
\end{table}

Table~\ref{tab:result_main_lade} reports the single-city results. Across all four cities, \textsc{Bridge} consistently improves over the baselines by a clear margin. Averaged over the four cities, retrieval augmentation reduces MAE from 3.245 to 3.152 and RMSE from 6.733 to 6.480 relative to the graph-only backbone. The gains are consistent: \textsc{Bridge} achieves the best MAE and RMSE across four datasets. The improvement is especially meaningful because the backbone already uses contextual region embeddings and inductive masking. This means the benefit does not come merely from adding static place semantics; instead, it comes from retrieving future patterns that complement the graph predictor when a region lacks local history. 
The largest RMSE reduction is observed in Hangzhou, while Chongqing shows the largest relative MAE improvement, suggesting that retrieval is especially effective in cities with recurring short-term demand dynamics that can be matched from memory.

\begin{figure*}[t]
    \centering
    \includegraphics[width=\textwidth]{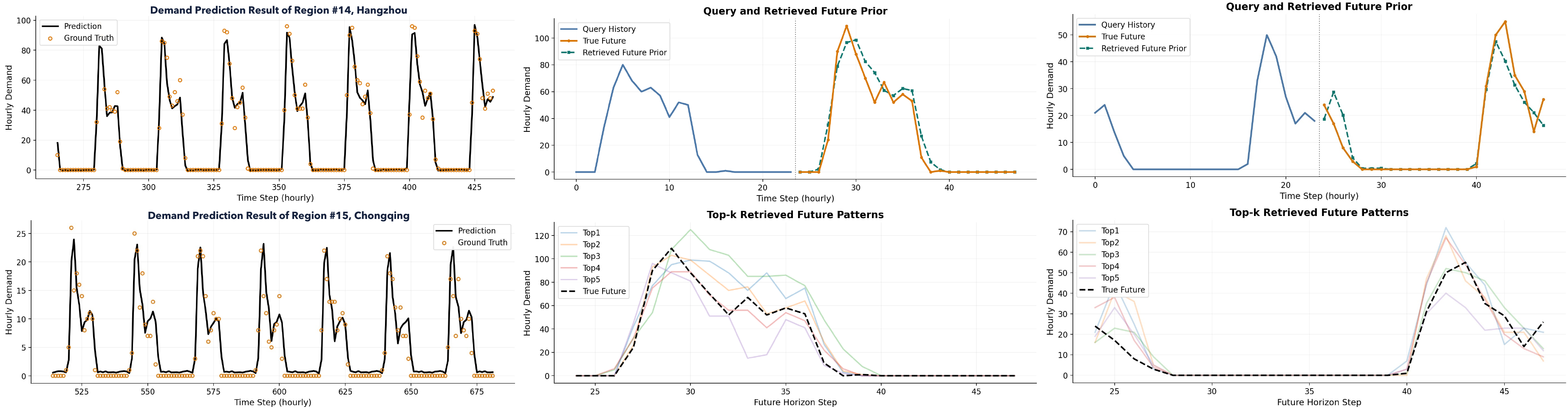}
    \vspace{-22pt}
    \caption{Qualitative forecasting examples. Left column: predicted and ground-truth demand curves for representative regions in Hangzhou and Chongqing. Middle column: retrieval behavior for a Shanghai single-city cold-start example, including the query history, retrieved future prior, and top-$k$ retrieved future patterns. Right column: retrieval behavior for a cross-city example, transferring from Yantai to Shanghai. In both cases, the retrieved trajectories capture the target temporal pattern and provide useful priors for the final fused forecast.}
    \label{fig:qualitative}
\vspace{-8pt}
\end{figure*}

\subsection{Qualitative Analysis}
Figure~\ref{fig:qualitative} provides qualitative examples from both settings. The left column shows node-level forecasting examples from Hangzhou and Chongqing, where \textsc{Bridge} tracks the repeated peaks and sparse intervals of the ground truth with high fidelity. The middle column shows a retrieval example from the Shanghai single-city cold-start setting. The right column shows a retrieval example when transferring from Yantai to Shanghai under partial target observations. In both retrieval examples, the retrieved future prior already matches the broad temporal pattern of the true future, including the timing of the major rise and the subsequent decay. The bottom panels further show that several top-$k$ retrieved futures share similar shapes, indicating that the retriever is not relying on a single brittle match but is aggregating a consistent set of candidate futures.
These visualizations illustrate the intended role of retrieval in \textsc{Bridge}. The graph backbone supplies a stable parametric forecast, while retrieval contributes an operational prior that captures short-lived patterns such as peak timing and amplitude changes. When the retrieved futures are well aligned with the target pattern, the fused prediction becomes noticeably sharper than what a graph-only predictor would produce from masked histories alone.

\section{Conclusion}
We presented \textsc{Bridge}, a retrieval-augmented spatiotemporal graph framework for cold-start urban delivery demand forecasting. The key idea is to complement inductive graph forecasting with a memory of delivery patterns, so the model can recover short-term operational dynamics when local histories are missing or only partially observed. Across two practical cold-start settings, \textsc{Bridge} consistently improves over the graph-only backbone and strong spatiotemporal baselines. These results indicate that future-prior retrieval is a practical way to strengthen delivery demand forecasting under realistic cold-start deployment conditions.

\bibliographystyle{IEEEtran}
\bibliography{ref-extracts}

@article{kipf2016semi,
  title={Semi-supervised classification with graph convolutional networks},
  author={Kipf, Thomas N and Welling, Max},
  journal={arXiv preprint arXiv:1609.02907},
  year={2016}
}

@inproceedings{tang2022domain,
  title={Domain adversarial spatial-temporal network: A transferable framework for short-term traffic forecasting across cities},
  author={Tang, Yihong and Qu, Ao and Chow, Andy HF and Lam, William HK and Wong, Sze Chun and Ma, Wei},
  booktitle={Proceedings of the 31st ACM international conference on information \& knowledge management},
  pages={1905--1915},
  year={2022}
}

@article{nie2025joint,
  title={Joint estimation and prediction of city-wide delivery demand: A large language model empowered graph-based learning approach},
  author={Nie, Tong and He, Junlin and Mei, Yuewen and Qin, Guoyang and Li, Guilong and Sun, Jian and Ma, Wei},
  journal={Transportation Research Part E: Logistics and Transportation Review},
  volume={197},
  pages={104075},
  year={2025},
  publisher={Elsevier}
}

@article{DCRNN,
  title={Diffusion convolutional recurrent neural network: Data-driven traffic forecasting},
  author={Li, Yaguang and Yu, Rose and Shahabi, Cyrus and Liu, Yan},
  journal={arXiv preprint arXiv:1707.01926},
  year={2017}
}

@article{STGCN,
  title={Spatio-temporal graph convolutional networks: A deep learning framework for traffic forecasting},
  author={Yu, Bing and Yin, Haoteng and Zhu, Zhanxing},
  journal={arXiv preprint arXiv:1709.04875},
  year={2017}
}

@article{GWNET,
  title={Graph wavenet for deep spatial-temporal graph modeling},
  author={Wu, Zonghan and Pan, Shirui and Long, Guodong and Jiang, Jing and Zhang, Chengqi},
  journal={arXiv preprint arXiv:1906.00121},
  year={2019}
}

@inproceedings{MTGNN,
  title={Connecting the dots: Multivariate time series forecasting with graph neural networks},
  author={Wu, Zonghan and Pan, Shirui and Long, Guodong and Jiang, Jing and Chang, Xiaojun and Zhang, Chengqi},
  booktitle={Proceedings of the 26th ACM SIGKDD international conference on knowledge discovery \& data mining},
  pages={753--763},
  year={2020}
}

@inproceedings{IGNNK,
  title={Inductive graph neural networks for spatiotemporal kriging},
  author={Wu, Yuankai and Zhuang, Dingyi and Labbe, Aurelie and Sun, Lijun},
  booktitle={Proceedings of the AAAI Conference on Artificial Intelligence},
  volume={35},
  number={5},
  pages={4478--4485},
  year={2021}
}

@article{SATCN,
  title={Spatial aggregation and temporal convolution networks for real-time kriging},
  author={Wu, Yuankai and Zhuang, Dingyi and Lei, Mengying and Labbe, Aurelie and Sun, Lijun},
  journal={arXiv preprint arXiv:2109.12144},
  year={2021}
}

@inproceedings{MPGRU,
  title={On the equivalence between temporal and static equivariant graph representations},
  author={Gao, Jianfei and Ribeiro, Bruno},
  booktitle={International Conference on Machine Learning},
  pages={7052--7076},
  year={2022},
  organization={PMLR}
}

@article{GRIN,
  title={Filling the g\_ap\_s: Multivariate time series imputation by graph neural networks},
  author={Cini, Andrea and Marisca, Ivan and Alippi, Cesare},
  journal={arXiv preprint arXiv:2108.00298},
  year={2021}
}

@article{gurnee2023language,
  title={Language models represent space and time},
  author={Gurnee, Wes and Tegmark, Max},
  journal={arXiv preprint arXiv:2310.02207},
  year={2023}
}

@article{manvi2023geollm,
  title={Geollm: Extracting geospatial knowledge from large language models},
  author={Manvi, Rohin and Khanna, Samar and Mai, Gengchen and Burke, Marshall and Lobell, David and Ermon, Stefano},
  journal={arXiv preprint arXiv:2310.06213},
  year={2023}
}

@article{cini2023graph,
  title={Graph deep learning for time series forecasting},
  author={Cini, Andrea and Marisca, Ivan and Zambon, Daniele and Alippi, Cesare},
  journal={arXiv preprint arXiv:2310.15978},
  year={2023}
}

@String{Computing = "Computing" }

@ArtifactSoftware{R,
    title = {R: A Language and Environment for Statistical Computing},
    author = {{R Core Team}},
    organization = {R Foundation for Statistical Computing},
    address = {Vienna, Austria},
    year = {2019},
    url = {https://www.R-project.org/},
}

@article{wu2023lade,
  title={Lade: The first comprehensive last-mile delivery dataset from industry},
  author={Wu, Lixia and Wen, Haomin and Hu, Haoyuan and Mao, Xiaowei and Xia, Yutong and Shan, Ergang and Zhen, Jianbin and Lou, Junhong and Liang, Yuxuan and Yang, Liuqing and others},
  journal={arXiv preprint arXiv:2306.10675},
  year={2023}
}

@article{hess2021real,
  title={Real-time demand forecasting for an urban delivery platform},
  author={Hess, Alexander and Spinler, Stefan and Winkenbach, Matthias},
  journal={Transportation Research Part E: Logistics and Transportation Review},
  volume={145},
  pages={102147},
  year={2021},
  publisher={Elsevier}
}

@article{liang2023poisson,
  title={A Poisson-based distribution learning framework for short-term prediction of food delivery demand ranges},
  author={Liang, Jian and Ke, Jintao and Wang, Hai and Ye, Hongbo and Tang, Jinjun},
  journal={IEEE Transactions on Intelligent Transportation Systems},
  year={2023},
  publisher={IEEE}
}

@article{srinivas2022autonomous,
  title={Autonomous robot-driven deliveries: A review of recent developments and future directions},
  author={Srinivas, Sharan and Ramachandiran, Surya and Rajendran, Suchithra},
  journal={Transportation research part E: logistics and transportation review},
  volume={165},
  pages={102834},
  year={2022},
  publisher={Elsevier}
}

@article{yannis2006effects,
  title={Effects of urban delivery restrictions on traffic movements},
  author={Yannis, George and Golias, John and Antoniou, Constantinos},
  journal={Transportation Planning and Technology},
  volume={29},
  number={4},
  pages={295--311},
  year={2006},
  publisher={Taylor \& Francis}
}

@article{lewis2020retrieval,
  title={Retrieval-augmented generation for knowledge-intensive nlp tasks},
  author={Lewis, Patrick and Perez, Ethan and Piktus, Aleksandra and Petroni, Fabio and Karpukhin, Vladimir and Goyal, Naman and K{\"u}ttler, Heinrich and Lewis, Mike and Yih, Wen-tau and Rockt{\"a}schel, Tim and others},
  journal={Advances in Neural Information Processing Systems},
  volume={33},
  pages={9459--9474},
  year={2020}
}

@article{wen2024survey,
  title={A survey on service route and time prediction in instant delivery: Taxonomy, progress, and prospects},
  author={Wen, Haomin and Lin, Youfang and Wu, Lixia and Mao, Xiaowei and Cai, Tianyue and Hou, Yunfeng and Guo, Shengnan and Liang, Yuxuan and Jin, Guangyin and Zhao, Yiji and others},
  journal={IEEE Transactions on Knowledge and Data Engineering},
  year={2024},
  publisher={IEEE}
}

@article{wei2024inductive,
  title={Inductive and adaptive graph convolution networks equipped with constraint task for spatial--temporal traffic data kriging},
  author={Wei, Tonglong and Lin, Youfang and Guo, Shengnan and Lin, Yan and Zhao, Yiji and Jin, Xiyuan and Wu, Zhihao and Wan, Huaiyu},
  journal={Knowledge-Based Systems},
  volume={284},
  pages={111325},
  year={2024},
  publisher={Elsevier}
}

@inproceedings{he2025geolocation,
  title={Geolocation representation from large language models are generic enhancers for spatio-temporal learning},
  author={He, Junlin and Nie, Tong and Ma, Wei},
  booktitle={Proceedings of the AAAI Conference on Artificial Intelligence},
  volume={39},
  number={16},
  pages={17094--17104},
  year={2025}
}

\end{document}